\documentclass[letterpaper, 10 pt, conference]{ieeeconf}  

\IEEEoverridecommandlockouts                              

\usepackage{graphicx}
\usepackage[
    backend=biber,
    style=ieee,
    sorting=none,
    mincitenames=1,
    maxcitenames=2,
    natbib=true,
]{biblatex} 
\usepackage{subcaption}
\DeclareCaptionLabelSeparator{periodspace}{.\quad}
\usepackage[usenames,dvipsnames]{color}
\usepackage{multirow}
\usepackage{tabularx}
\usepackage{bm}
\usepackage{amsmath}
\usepackage{amssymb}
\usepackage[hidelinks]{hyperref}
\usepackage{siunitx}
\usepackage[capitalise]{cleveref}

\usepackage{xcolor}
\colorlet{mygreen}{black!50!green}
\definecolor{urlcolor}{RGB}{23, 17, 150}
\newcommand*{\draftTikz}{}
\ifdefined\draftTikz
    \usepackage{tikz}
    \usetikzlibrary{arrows,calc}
    \usepackage{pgfplots}
    \pgfplotsset{compat=newest}
    \usetikzlibrary{external}
    \tikzset{%
        font=\footnotesize,
        external/up to date check=md5,
    }
\fi

\title{\LARGE \bf
Nonprehensile Planar Manipulation through Reinforcement Learning with Multimodal Categorical Exploration
}

\author{Juan Del Aguila Ferrandis$^{1}$, Jo\~{a}o Moura$^{1,2}$, and Sethu Vijayakumar$^{1,2}$
\thanks{
$^{1}$School of Informatics, The University of Edinburgh, Edinburgh, UK.
$^{2}$The Alan Turing Institute, London, U.K.
This research is supported by the EU H2020 projects Enhancing Healthcare with Assistive Robotic Mobile Manipulation (HARMONY, 101017008) and the Kawada Robotics Corporation.
}
}

\begin{document}

\maketitle
\thispagestyle{empty}
\pagestyle{empty}

\begin{abstract}

Developing robot controllers capable of achieving dexterous nonprehensile manipulation, such as pushing an object on a table, is challenging.
The underactuated and hybrid-dynamics nature of the problem, further complicated by the uncertainty resulting from the frictional interactions, requires sophisticated control behaviors.
Reinforcement Learning (RL) is a powerful framework for developing such robot controllers. 
However, previous RL literature addressing the nonprehensile pushing task achieves low accuracy, non-smooth trajectories, and only simple motions, i.e. without rotation of the manipulated object.
We conjecture that previously used unimodal exploration strategies fail to capture the inherent hybrid-dynamics of the task, arising from the different possible contact interaction modes between the robot and the object, such as sticking, sliding, and separation.
In this work, we propose a multimodal exploration approach through categorical distributions, which enables us to train planar pushing RL policies for arbitrary starting and target object poses, i.e. positions and orientations, and with improved accuracy.
We show that the learned policies are robust to external disturbances and observation noise, and scale to tasks with multiple pushers.
Furthermore, we validate the transferability of the learned policies, trained entirely in simulation, to a physical robot hardware using the KUKA iiwa robot arm.
See our supplemental video: \href{https://youtu.be/vTdva1mgrk4}{\textcolor{urlcolor}{https://youtu.be/vTdva1mgrk4}}.


\end{abstract}

\section{Introduction}

Nonprehensile manipulation, defined as manipulation without grasping, endows robots with versatile behaviors, enabling them to perform a wide range of motions on objects with different properties \cite{mason_1986, mason_1999}. 
However, allowing the pose of the object relative to the end-effector to change requires the robot to constantly adapt the contact positions, leading to different possible contact modes in the form of sticking, sliding, and separation. 
As a result, multiple interesting challenges arise. 
Most notably, the underactuated nature of the system makes it infeasible to realize arbitrary motions of the object \cite{hogan_2020}, in addition to the complexity of hybrid-dynamics resulting from the transitions between different contact modes \cite{hogan_2020}, and the hard to model frictional interactions exacerbating the uncertainty in the contact modes and the object motion \cite{zhou_2017, bauza_2017}. 

\begin{figure}[tp]
    \centering
    \includegraphics[width=\linewidth]{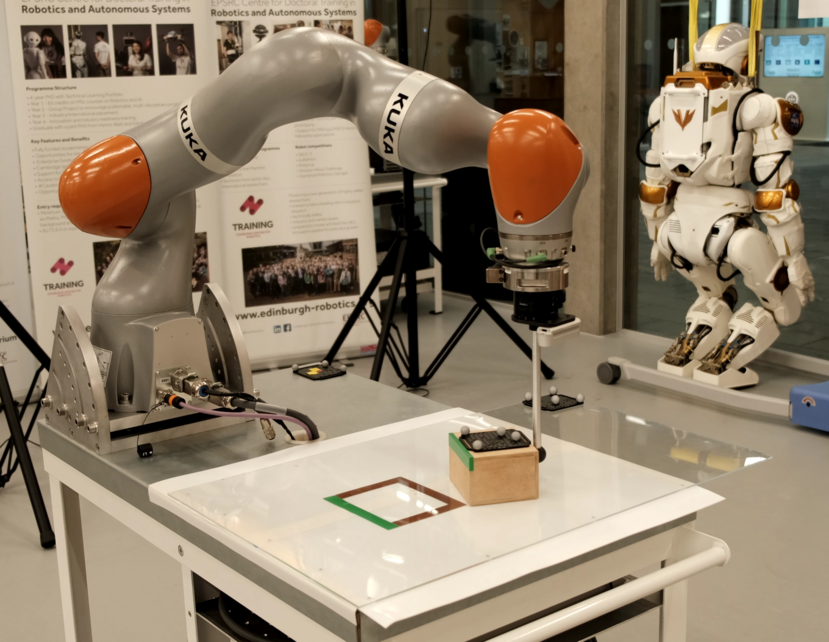}
    \caption{Experimental robotic hardware set-up for the planar pushing task. The robot uses a pusher to move an object to a specified target pose.}
    \label{fig:robot_picture}
\end{figure}

In this paper we consider the task of planar pushing, widely studied in the nonprehensile literature \cite{mason_1986, hogan_2020, bauza_2017, goyal_1989, moura_2022}. 
The task, as seen in~\cref{fig:robot_picture}, consists of using a robotic pusher to control the motion of an object sliding on a flat surface. 
Previous works developed robot controllers for planar pushing, generally following one of two approaches: model-based via Model Predictive Control (MPC) \cite{hogan_2020, moura_2022}, or model-free via Reinforcement Learning (RL) \cite{peng_2018, lowrey_2018, cong_2022, jeong_2019}. 
These approaches typically face different open problems; MPC lacks scalability to more complex scenarios, such as multiple contacts and switching contact faces \cite{hogan_2020, xue_2022}, while RL methods tend to produce non-smooth robot motions and, in the case of planar pushing, show limited range of sub-optimal, idiosyncratic motions, which will be the focus of this paper. 

\subsection{Related Work}
Mason's \cite{mason_1986} seminal work introduced planar pushing as a simple task to study broader concepts in modeling, planning, and control.
This work developed  analytical models of the dynamics of planar pushing by exploring the concepts of the voting theorem and the motion cone.
Later on,~\citet{goyal_1989} added the concept of the limit surface to model the friction between the object and the table.
More recently, \citet{zhou_2016} and \citet{bauza_2017} explored data-driven approaches for modeling the dynamics of planar pushing through convex polynomials and Gaussian processes respectively.

\citet{hogan_2020} proposed a mixed integer MPC formulation for offline tracking of nominal trajectories, enabling the use of sticking and sliding contact modes for pushing, and learned an approximation of the contact mode selection for an online deployment of the MPC. 
\citet{moura_2022} introduced a complementarity constraint MPC formulation for online tracking of nominal trajectories, i.e.~without pre-learning the contact mode selection.
This formulation also enables offline trajectory optimization with sticking and sliding contact modes.
Both of these approaches achieve the target poses with significant accuracy while recovering from disturbances, however, require offline generation of nominal trajectories and 
rely on the quasi-static assumption, which limits the pushing velocities below \SI{0.08}{\meter\per\second}~\cite{bauza_2017}.

RL methods can overcome some of the online scalability limitations of MPC through extensive offline exploration during training. 
\citet{peng_2018} used dynamics randomization to transfer configuration-space planar pushing RL policies from simulation to the physical robotic set-up. 
\citet{lowrey_2018} also explored RL for planar pushing using a different robotic set-up, consisting of 3 pushers, each with 3 degrees of freedom (DOF).
More recently,~\citet{cong_2022} used RL with vision-proprioception to learn task-space planar pushing motions for objects with different shapes. 
These RL methods learn a policy suitable for online execution and can recover from significant disturbances. 
However, they result in non-smooth trajectories with overall poor accuracy, i.e.~position errors greater than~\SI{2}{\cm}, and crucially, they disregard the orientation of the object. 

\subsection{Problem Statement}

While model-free RL methods address the scalability limitations of model-based approaches, for example with respect to the number of contacts and ability to switch contact faces, the current literature on the application of RL methods to planar pushing achieve low accuracy, non-smooth trajectories, and only simple motions, i.e. without orientation of the sliding object
~\cite{peng_2018,lowrey_2018,cong_2022,jeong_2019}, which we aim to consider.
The aforementioned RL methods share a common trait: they formulate the task with a continuous action space and use a multivariate Gaussian with diagonal covariance for exploration. 
This limits the exploration to unimodal policies across each action space dimension. 
However, the model-based literature identifies the planar pushing problem as a hybrid-dynamic system due to the different possible contact modes (sticking, sliding left, sliding right, and separation).
This provides us with the insight that perhaps planar pushing is fundamentally a multimodal control problem.
Therefore, we ask the question:~\textit{can multimodal exploration enable us to learn robust, scalable, and accurate planar pushing RL policies that incorporate object orientation?}

\subsection{Contributions}

In this paper we make the following contributions:
\begin{itemize}
    \item We propose a multimodal exploration approach, with categorical distributions on a discrete action space, which enables us to learn planar pushing RL policies for arbitrary initial and target object poses, i.e.~different positions and orientations.
    \item We demonstrate that the proposed framework is robust to disturbances and observation noise, scalable to two pushers, and exhibits smooth pushing motions.
    \item We validate the policies, trained only in simulation, on a physical hardware set-up using the KUKA iiwa robot.
\end{itemize}

\section{Background}

\subsection{Planar Pushing}
\begin{figure}[ht]
    \centering
    \ifdefined\draftTikz
        \begin{tikzpicture}
    \def\U{1.1cm}
    \def\squared{1.0*\U}
    \def\halfx{0.6*\U}
    \def\halfy{0.9*\U}
    \def\angletarget{30}
    \def\anglestart{-30}
    \coordinate (origin) at (0,0);
    \coordinate (pusher) at (0.5*\U,1.3*\U);
    \coordinate (slider) at (1.7*\U,1.0*\U);
    \coordinate (boxtarget) at (4.7,0.9*\U);
    \draw[dashed] (slider) -- ++(1.0*\U,0);
    \draw[dashed, rotate=\anglestart] (slider) -- ++(1.0*\U,0);
    \draw[<->] ([shift=(0:0.9*\U)]slider) arc (0:\anglestart:0.9*\U) node[midway,right] {$\theta_b$};
    \shadedraw[shading angle=\anglestart, rotate=\anglestart, thick, rounded corners=0.5pt] ($(slider)+(\halfx,\halfy)$) rectangle ($(slider)+(-\halfx,-\halfy)$);
    \draw[-latex, black, thick, rotate=\anglestart] (slider) --++ (0.45*\U,0);
    \fill[black] (slider) circle (1.5pt) node[above] {$(x_b,y_b)$};
    \draw[dashed] (boxtarget) -- ++(1.0*\U,0);
    \draw[dashed, rotate=\angletarget] (boxtarget) -- ++(1.0*\U,0);
    \draw[<->] ([shift=(0:0.9*\U)]boxtarget) arc (0:\angletarget:0.9*\U) node[midway,right] {$\theta_{\text{targ}}$};
    \draw[mygreen, dashed, ultra thick, rotate=\angletarget, rounded corners=3pt] ($(boxtarget)+(\halfx,\halfy)$) rectangle ($(boxtarget)+(-\halfx,-\halfy)$);
    \draw[-latex, mygreen, very thick, rotate=\angletarget] (boxtarget) --++ (0.45*\U,0);
    \fill[black] (boxtarget) circle (1.5pt) node[below] {\hspace{5pt} $(x_{\text{targ}},y_{\text{targ}})$};
    \draw[thick, fill=blue] (pusher) circle circle (3pt) node[below left] {$(x_p,y_p)$};
    \draw[->, thick, blue] (pusher) -- ++(0.2*\U,0.6*\U) node[above] {$(v_{x,p},v_{y,p})$};
    \draw[thick,->] (origin) -- ++(0.5*\U,0) node[black,below] {$x$};
    \draw[thick,->] (origin) -- ++(0,0.5*\U) node[black,left] {$y$};
\end{tikzpicture}
    \else
        \includegraphics{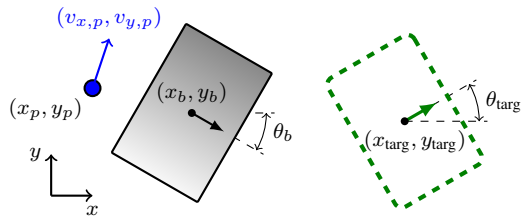}
    \fi
    \caption{Illustration of the planar pushing system.}
    \label{fig:box_illustration}
\end{figure}

We consider the task of pushing a box to a specified target pose, composed of the object position and orientation, from a random initial system configuration, composed of the initial object pose and robot pusher position, all within a bounded planar workspace. 
\cref{fig:box_illustration} illustrates the planar pushing system, where  $(v_{x,p}, v_{y,p})$ is the velocity of the pusher, located at $(x_p, y_p)$, $(x_b, y_b, \theta_b)$ is the pose of the box, and $(x_{\text{targ}}, y_{\text{targ}}, \theta_{\text{targ}})$ is the target pose.

\subsection{Problem Formulation}
We formulate the problem using goal-conditioned RL. 
In particular, we consider a finite horizon goal-conditioned Partially Observable Markov Decision Process (POMDP) defined by the tuple $(\mathcal{S}, \Omega, \mathcal{G}, \mathcal{A}, \mathcal{O}, \mathcal{P}, \mathcal{R}, H, \rho_0, \rho_g)$ \cite{cong_2022, miki_2022}. 
At each time step $t$, the environment has state $s_t \in \mathcal{S}$, we receive an observation $o_t \in \Omega$, our goal is $g_t \in \mathcal{G}$, and we take an action $a_t \in \mathcal{A}$. 
Note that the goal remains fixed during an episode. 
Additionally, $\mathcal{O} : \mathcal{S} \times \mathcal{A} \to \text{Pr}(\Omega)$ is the observation model, $\mathcal{P} : \mathcal {S} \times \mathcal{A} \to \text{Pr}(\mathcal{S})$ is the transition dynamics, and $\mathcal{R} : \mathcal{S} \times \mathcal{A} \times \mathcal{G} \to \mathbb{R}$ is the reward function. 
We limit episodes to have a maximum horizon $H$. 
Finally, the initial state and goal of an episode are distributed according to $\rho_0$ and $\rho_g$ respectively. 

\subsection{Proximal Policy Optimization}
We wish to learn a stochastic policy $\pi_{\theta} : \Omega \times \mathcal{G} \to \text{Pr}({\mathcal{A}})$ parametrized by $\theta$. 
To this end, we use Proximal Policy Optimization (PPO) \cite{schulman_2017}, which is a popular on-policy RL algorithm widely applied in various control tasks, including in-hand manipulation \cite{openai_2018} and locomotion \cite{deepmind_2017}. 
PPO uses a truncated Generalized Advantage Estimation (GAE) \cite{schulman_gae_2016} to estimate the advantage function for time step $t \in [0, T]$ as
\begin{equation}
    \hat{A}_t = \sum_{i = 0}^{T-t} (\gamma\lambda)^{i}\delta_{t+i},
\end{equation}
where
\begin{equation}
    \delta_t = r_t + \gamma V_{{\phi}}({o}_{t+1}, {g}_{t+1}) - V_{{\phi}}({o}_t, {g}_t),
\end{equation}
$r_t$ is the reward, $V_{{\phi}} : \Omega \times \mathcal{G} \to \mathbb{R}$ is the value function, parametrized by $\phi$, $\lambda$ is the GAE parameter, and $\gamma$ is the discount factor.
Then, $\pi_{\theta}$ and $V_{\phi}$ can be learned together through mini-batch stochastic gradient ascent on the objective function
\begin{align}
L_t(\theta, \phi) = \ &\mathbb{\hat{E}}_{t} \bigg[ \min\left(\frac{\pi_{{\theta}}({a}_t \ \vert \ {o}_t, g_t)}{\pi_{{\theta}_{old}}({a}_t \ \vert \ {o}_t, g_t)} \cdot \hat{A}_t , \right. \nonumber\\ 
& \left. \text{clip}\left(\frac{\pi_{{\theta}}({a}_t \ \vert \ {o}_t, g_t)}{\pi_{{\theta}_{old}}({a}_t \ \vert \ {o}_t, g_t)}, \ 1 - \epsilon, \ 1 + \epsilon \right)\hat{A}_t \right) \nonumber\\ 
&  - c_1 L_t^{V_{{\phi}}} + c_2 S_t^{\pi_{{\theta}}}\bigg],
\end{align}
where we compute the expectation over a mini-batch of samples.
The first term within the expectation is the surrogate objective of the policy, $L_t^{V_{{\phi}}}$ is the loss of the value function, $S_t^{\pi_{{\theta}}}$ is an entropy bonus, $c_1, c_2$ are weights, and $\epsilon$ controls the clip range \cite{schulman_2017}.

\section{Method}

\subsection{Observation, Goal, Action, and Reward}

\textbf{Observation.}
At each time step $t$, the policy receives an observation $o_t$ consisting of  the current box pose $(x_b, y_b, \theta_b)$, and the current pusher position $(x_p, y_p)$. 
However, there is important information from the environment state $s_t$ that this observation fails to capture, for instance, the frictional contact forces and the velocity of the box.
We consider two techniques that attempt to enable the policy and value functions to capture this hidden information.
The first technique involves using a stack of previous observations $\{o_t, o_{t-1}, ...\ , o_{t-l}\}$ \cite{frame_stacking, peng_2018}. 
This allows us to use a simple architecture for the policy and value functions, such as a Multilayer Perceptron (MLP), which can extract an estimate of the hidden dynamics of the environment from the stack of observations.
The second technique involves using an LSTM layer within the policy and value functions \cite{peng_2018, openai_2018}.
This avoids the need for observation stacking by leveraging the hidden and cell states within the LSTM layer which have predictive capabilities of the environment dynamics.

\textbf{Goal and Action.}
The goal $g_t$ of the policy is to reach a particular target box pose $(x_{\text{targ}}, y_{\text{targ}}, \theta_{\text{targ}})$. 
Given a goal $g_t$ and an observation $o_t$, the policy takes an action $a_t = (v_{x,p}, v_{y,p})$, which consists of the $x$ and $y$ velocity of the pusher.
We limit the velocity on each axis to the range $[-0.1, 0.1]$~\SI{}{\meter\per\second}.

\textbf{Reward.}
If the object reaches the target, the episode terminates with a positive reward $r_t = \alpha$. 
Alternatively, if the object fails to reach the target within the maximum horizon, or the workspace boundaries are violated by the pusher or box, the episode terminates with a negative reward $r_t = -\beta$. 
Otherwise, the policy receives a reward $r_t = k_1 (1 - d_{x,y}) + k_2 (1 - d_\theta) + k_3 (1-v_p)$, where $d_{x,y}$ is the normalized distance to the target position, $d_\theta$ is the normalized angular distance to the target orientation, $v_p$ is the normalized magnitude of the pusher velocity, and $k_1, k_2, k_3$ control the weights of the three terms. 
The first two terms, corresponding to $k_1$ and $k_2$, provide signals reflecting the desirability of the current box pose relative to the target pose. 
The last term, corresponding to $k_3$, acts as a regularizer designed to encourage efficient motions of the pusher. 

\subsection{Exploration Strategies}
As discussed in the Problem Statement, previous works on RL for planar pushing perform exploration using a multivariate Gaussian with diagonal covariance. 
There are different possible formulations of this strategy, usually depending on the RL algorithm used.
When applying PPO with Gaussian exploration to our planar pushing set-up, the policy function outputs the mean velocities in $x$ and $y$, denoted as $\mu_x$ and $\mu_y$.
Combining them with the corresponding learned state-independent variances $\sigma_x^2$ and $\sigma_y^2$ results in a multivariate Gaussian from which we can sample the action \cite{schulman_2017}. 

Soft Actor Critic (SAC) \cite{haarnoja_2018} is a popular off-policy RL algorithm.
We include SAC with Gaussian exploration and an MLP architecture as a baseline in our experiments since the current state-of-the-art RL policies for planar pushing use the same configuration \cite{cong_2022}.
When using SAC with Gaussian exploration, the policy function directly outputs $\mu_x$, $\mu_y$, $\sigma_x^2$, and $\sigma_y^2$, which allows for the variances to be state-dependent \cite{haarnoja_2018}. 

A straightforward method to enable multimodal exploration strategies during policy training is to discretize the action space and use categorical distributions for exploration,  which can approximate any type of distribution. 
In particular, we discretize $v_{x,p}$ and $v_{y,p}$ using 11 bins for each velocity \cite{tang_2020, openai_2018}. 
Then, given an observation and goal pair, the policy network outputs 11 logits that define a categorical distribution over $v_{x,p}$ and another 11 logits that define a categorical distribution over $v_{y,p}$.
We sample the action from these distributions.

\subsection{Sim-to-Real Transfer}
We train the policies entirely in simulation and use dynamics randomization, observation noise, and synthetic disturbances to bridge the sim-to-real gap. 
At the start of every episode, we sample random values for: (a) the friction and restitution of the floor, box, and pusher; (b) the dimensions of the box and the pusher; and (c) the mass of the box. 
Additionally, at every time step we randomize the time duration of the action \cite{peng_2018}.
We also add correlated noise, sampled at the beginning of each episode, and uncorrelated noise, sampled at every time step, to the observations of the box pose and pusher position, to simulate sensor uncertainty. 
Finally, we apply random disturbances to the box during each training episode. 

\subsection{Curriculum Learning}
\label{sec:curriculum}
The goal of the policy is to move the box to the target pose.
Therefore, we define thresholds $T_{x,y}$ and $T_\theta$, corresponding to the position and the orientation respectively, such that, if $\|(x_b, y_b) - (x_{\text{targ}}, y_{\text{targ}})\| \leq T_{x,y}$ and $\vert \theta_b - \theta_{\text{targ}}\vert \leq T_\theta$, then the episode terminates as a success.
Smaller $T_{x,y}$ and $T_\theta$ lead to more accurate learned policies; however, this is at the expense of increased task complexity and a sparser reward signal, which can lead to much slower learning, or lack of convergence entirely.
To mitigate this issue, we define a curriculum such that the learning starts with larger thresholds $T_{x,y}, T_\theta$, and they are reduced to $T_{x,y}/2$ and $T_\theta/2$ if the policy reaches a 90\% average success rate.

\section{Experimental Set-up}

\textbf{Training.}
We train the policies with data collected from 128 parallel actors in simulated planar pushing systems. 
The simulations are performed using PyBullet \cite{coumans2021} and, by default, the policies run at a frequency of 30 Hz.
Additionally, the maximum episode length is $H = 300$ time steps, corresponding to 10 seconds in real-time. 
For the reward function, we use parameter values $\alpha = 50$, $\beta = 20$, $k_1 = 0.1$, $k_2 = 0.02$, and $k_3 = 0.004$.
Our implementations of PPO and SAC are based on Stable Baselines3 \cite{raffin_2021}.
We design a custom planar pushing environment for learning.
All policies are trained in a single workstation with an Intel Core i9 3.60GHz, GeForce RTX 2080, and 64 GiB of RAM.

\textbf{Network Architecture.}
We experiment with two different neural network architectures for the policy and value functions in PPO. 
The first architecture consists of an MLP which receives as an input the goal $g_t$ and a stack of the 10 previous observations $\{o_t, o_{t-1}, ...\ , o_{t-9}\}$. 
The policy function contains 2 hidden linear layers, each of size 512, while the value function contains 2 hidden linear layers, each of size 1024. 
The second architecture involves an LSTM layer, which allows us to include only the goal $g_t$ and the current observation $o_t$ in the input.
In this case, the policy and value functions have the same shape and contain the following hidden layers: a linear layer of size 128, an LSTM layer of size 256, and a linear layer of size 128, in this order.  
We use $\tanh$ nonlinearities in both architectures. 

\textbf{PPO Hyperparameters.}
During training, excessively large model updates can lead to policy collapse. 
One way to mitigate this in PPO is through early stopping of model updates when the KL divergence of the new policy and the old policy exceeds a certain threshold \cite{raffin_2021}. 
We use such early stopping with a threshold of 0.01 when training our policies.
The remaining hyperparameters used in PPO are summarized in \cref{ppo_hyperparameters}.

\begin{table}[t]
\caption{Hyperparameters for PPO}
\label{ppo_hyperparameters}
\centering
{\renewcommand{\arraystretch}{1.2}
\begin{tabular}{l|l}
\hline
\textbf{Hyperparameter}      & \textbf{Value} \\ \hline
Clip range ($\epsilon$)      & 0.2               \\
GAE parameter ($\lambda$)    & 0.95               \\
Discount factor ($\gamma$)   & 0.99               \\
Value function coefficient ($c_1$)  & 0.5  \\
Entropy bonus coefficient ($c_2$) & 0 \\
Epochs                       & 10  \\
Optimizer                    & Adam \cite{adam}              \\
Learning rate                & $3\cdot10^{-4}$               \\
Batch size                   & 7680               \\
\end{tabular}
}
\end{table}

\textbf{SAC Hyperparameters.}
For our experiments with SAC, we use the same MLP architecture as in PPO with minor modifications since SAC learns two state-action value functions instead of a single state value function, and applies $\operatorname{ReLu}$ instead of $\tanh$ nonlinearities \cite{haarnoja_2018}.
Additionally, we use the same optimizer, batch size, and learning rate as in PPO.
SAC is an off-policy algorithm and hence stores previous trajectories in a replay buffer. 
We use a buffer of size $10^{6}$.
Finally, we apply a $\tanh$ squashing function to the action sampled from the policy \cite{haarnoja_2018,cong_2022, raffin_2021}.

\textbf{Starting State and Goal.}
At the beginning of every episode, we generate a random starting configuration and a random target box pose. 
The starting and target box positions are independently and uniformly sampled from the available workspace.
Additionally, the starting and target box orientations are independently sampled from $\mathcal{U}([-\pi, \pi])$ \SI{}{\radian}. 
The starting pusher position is sampled uniformly from a perimeter around the box. 
This is done to facilitate initial exploration and, as can be seen in the supplemental video, it does not prevent the policy from learning how to reach the box from large separations, presumably due to the synthetic disturbances during training.

\textbf{Episode Success.}
During the first stage of the training process, we use success thresholds $T_{x,y} = \SI{1.5}{\cm}$ and $T_\theta = \SI{0.34}{\radian} \approx 19.5$°. Then, if the policy exceeds a 90\% success rate, averaged over the last 100 episodes completed by each of the 128 parallel actors, we halve the success thresholds to $T_{x,y} = \SI{ 0.75}{\cm}$ and $T_\theta = \SI{0.17}{\radian} \approx 9.7$°. An episode is successful if the box reaches the target pose within the current success thresholds and the box has velocity \SI{0}{\meter\per\second}. On the other hand, an episode is unsuccessful if the maximum episode length is reached without achieving the goal, as well as if the pusher or the box leave the workspace. 

\begin{table}[t]
\caption{Dynamics Randomization and Observation Noise Parameters}
\label{randomization_parameters}
\centering
{\renewcommand{\arraystretch}{1.2}
\begin{tabular}{l|l}
\hline
\textbf{Parameter} & \textbf{Sampling Distribution}\\
\hline
Friction & $\mathcal{U}([0.5, 0.7])$\\
Restitution & $\mathcal{U}([0.4, 0.6])$\\
Box Length & $\mathcal{U}([0.115, 0.125])$ \SI{}{\meter}\\
Box Width & $\mathcal{U}([0.095, 0.105])$ \SI{}{\meter}\\
Box Mass & $\mathcal{U}([0.4, 0.6])$ \SI{}{\kilogram}\\
Pusher Radius & $\mathcal{U}([0.012, 0.013])$ \SI{}{\meter}\\
Action Duration & $\mathcal{N}(1/30, (1/320)^2)$ \SI{}{\second}\\
Position Noise & $\mathcal{N}(0, 0.001^2)$ \SI{}{\meter}\\
Orientation Noise & $\mathcal{N}(0, 0.02^2)$ \SI{}{\radian}\\
\end{tabular}
}
\end{table}

\textbf{Randomization Parameters.}
\cref{randomization_parameters} details the parameters and corresponding sampling distributions for the dynamics randomization and observation noise. 
We sample the dynamics parameters from their respective distributions at the beginning of each episode. 
Additionally, we independently sample correlated and uncorrelated observation noise for the position of the box $(x_b, y_b)$ and the position of the pusher $(x_p, y_p)$ from the Position Noise distribution in Table~\ref{randomization_parameters}. 
For example, let $\mathcal{O}(x_b)$ denote the policy observation of $x_b$. 
Then, during training, $\mathcal{O}(x_b) = x_b + \delta_{x_b}^{\text{episode}} + \delta_{x_b}^{\text{step}}$, where $\delta_{x_b}^{\text{episode}}$ is sampled at the beginning of the episode, and $\delta_{x_b}^{\text{step}}$ is sampled at every time step. 
Similarly, we add correlated and uncorrelated noise to the observation of the box orientation $\theta_b$ by sampling from the Orientation Noise distribution in \cref{randomization_parameters}. 
Finally, at every time step we apply a disturbance to the box with probability 1\%, in a uniformly random position, and with force in $x$ and $y$ independently sampled from $\mathcal{U}([-25, 25])$ \SI{}{\newton}.

\textbf{Two-pusher Set-up.}
We also conduct experiments using two pushers to evaluate the scalability of our framework.
This requires certain adjustments to the training procedure. 
We augment the action space to include the $x$ and $y$ velocity of the two pushers.
Additionally, in order to guarantee that the policy only explores motions that are feasible for a bi-manual manipulation platform, we add two constraints: (a) each pusher can exert pushing forces with a maximum magnitude of \SI{75}{\newton}; and (b) the distance between pushers in the $x$ coordinate must be at least \SI{5}{\cm}.
If the policy violates any of these constraints, the episode terminates unsuccessfully.

\section{Experiments and Results -- Simulation}

\begin{figure}[t]
    \centering
    \includegraphics[trim={0 0.2cm 0 1.15cm},clip,width=\linewidth]{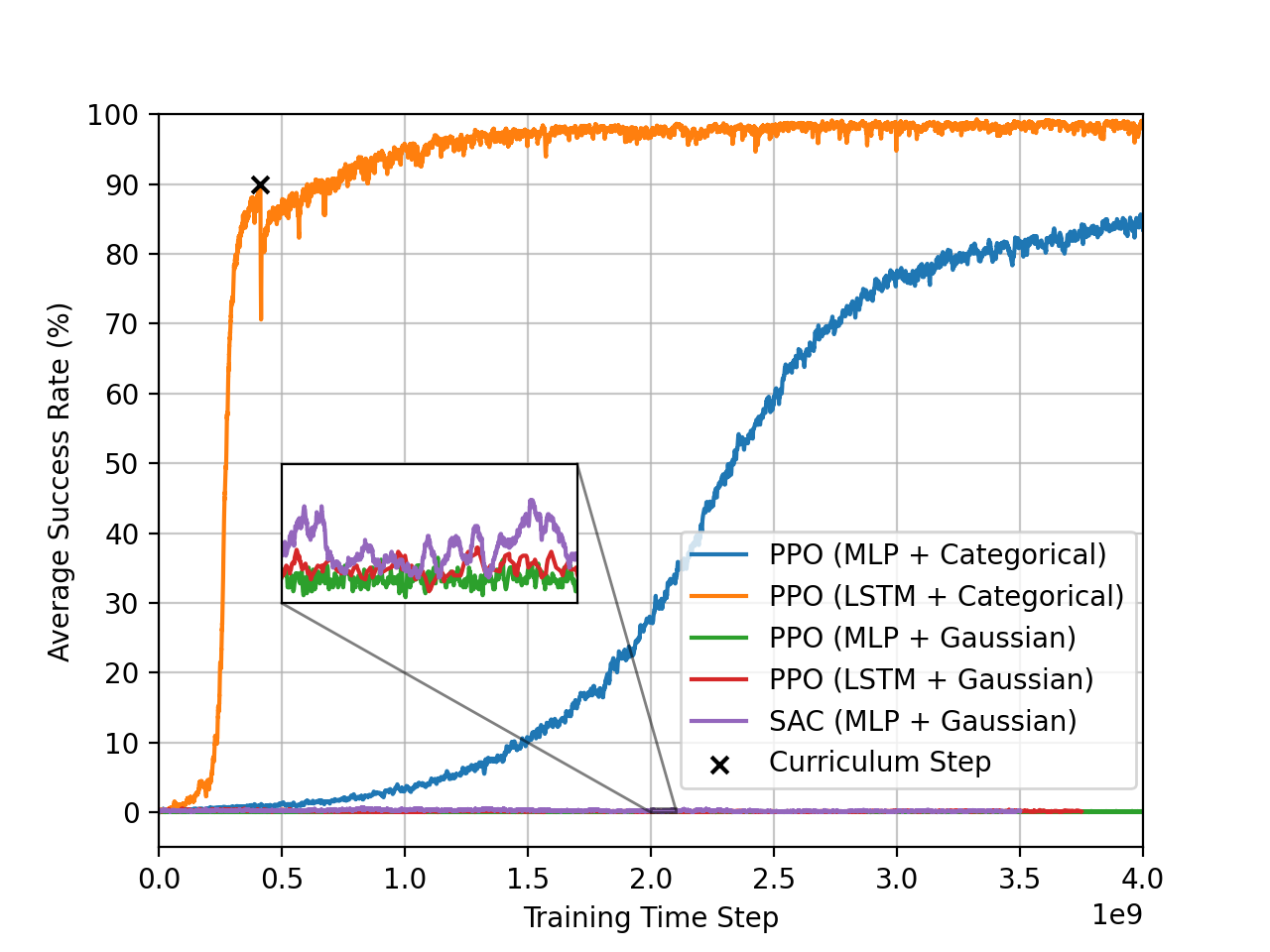}
    \caption{Training performance of the evaluated frameworks. Success rate is averaged over the last 100 episodes completed by each of the parallel actors. The curriculum halves the success thresholds at 90\% average success rate.}
    \label{fig:training_curves_one_pusher}
\end{figure}

\begin{figure}[b]
    \centering
    \includegraphics[width=\linewidth]{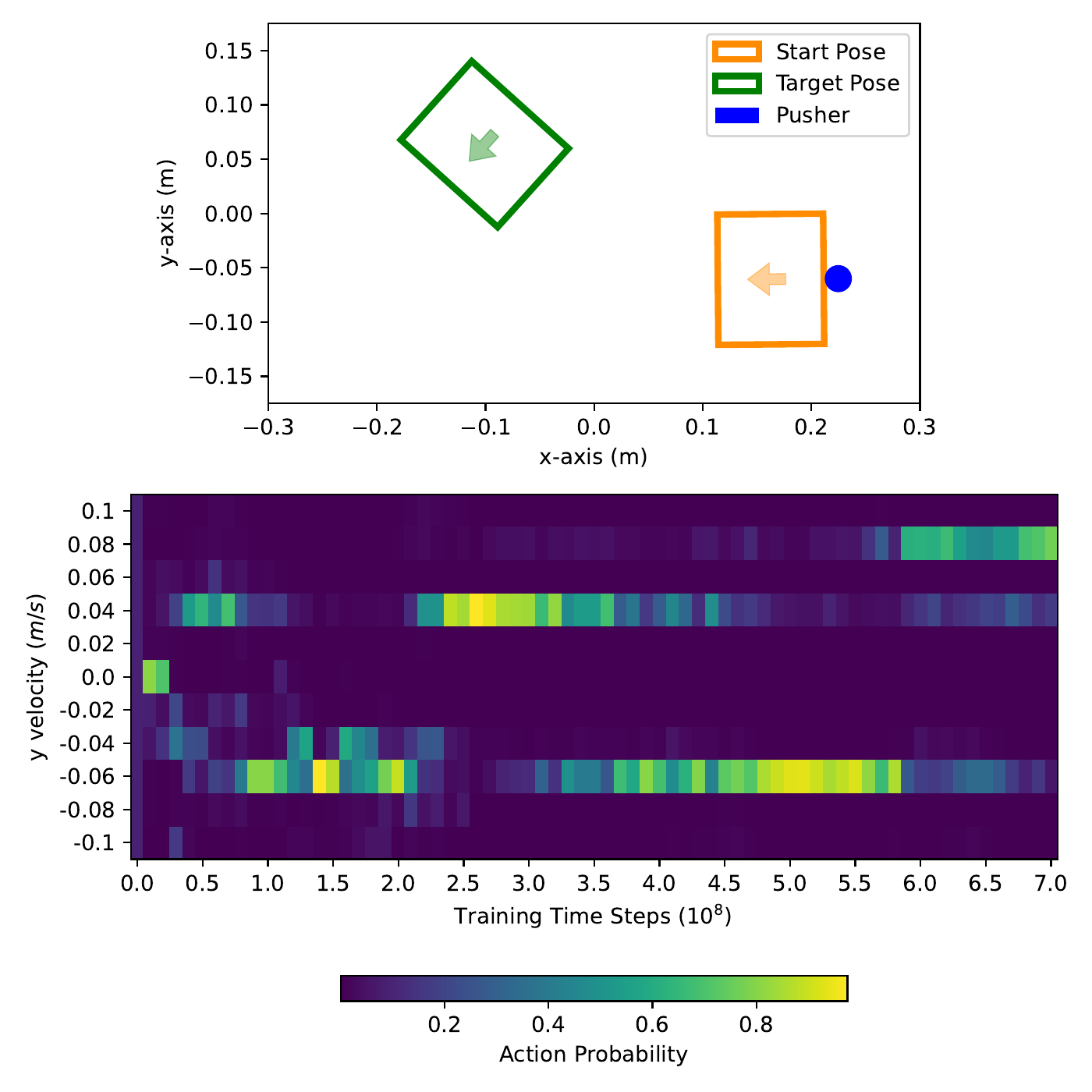}
    \caption{Evolution during training of the categorical action distribution for the pusher velocity in the $y$ axis $\left(v_{y,p}\right)$ for the configuration shown above.}
    \label{fig:action_distribution}
\end{figure}

We first consider the standard set-up with one pusher.
We train PPO policies with the MLP and LSTM architectures, and compare the categorical and Gaussian exploration strategies for each configuration.
We also train a SAC policy with the MLP architecture and Gaussian exploration. 
The resulting learning curves are shown in Fig~\ref{fig:training_curves_one_pusher}.
We find that only the policies that use the proposed categorical exploration approach manage to learn the task. 
Additionally, the LSTM architecture provides substantially faster convergence.

The best policy is PPO (LSTM + Categorical).
It converges rapidly and reaches a 90\% average success rate early on, which triggers the curriculum step where $T_{x,y}$ and $T_\theta$ are halved.
Hence, the policy manages to learn significantly accurate planar pushing motions. 
In particular, during training, the policy achieves over 98\% average success rate with success thresholds $T_{x,y} = \SI{0.75}{\cm}$ and $T_\theta = \SI{0.17}{\radian} \approx 9.7$°.


We further investigate whether exploration through categorical distributions indeed leads to multimodal strategies.
We examine the evolution during training of the categorical distribution in PPO (LSTM + Categorical) for the action $v_{y,p}$ in various environment states.
As expected, we often find that the action distribution is multimodal.
\cref{fig:action_distribution} shows the results for one of these cases.
In particular, it broadly has two modes that correspond to upward and downward motions.
Therefore, it seems that the categorical exploration strategy enables the policy to explore different possible contact modes concurrently during training. 

\begin{figure}[t]
    \begin{subfigure}{\linewidth}
        \centering
        \includegraphics[trim={0 1.2cm 0 2.2cm},clip,width=\linewidth]{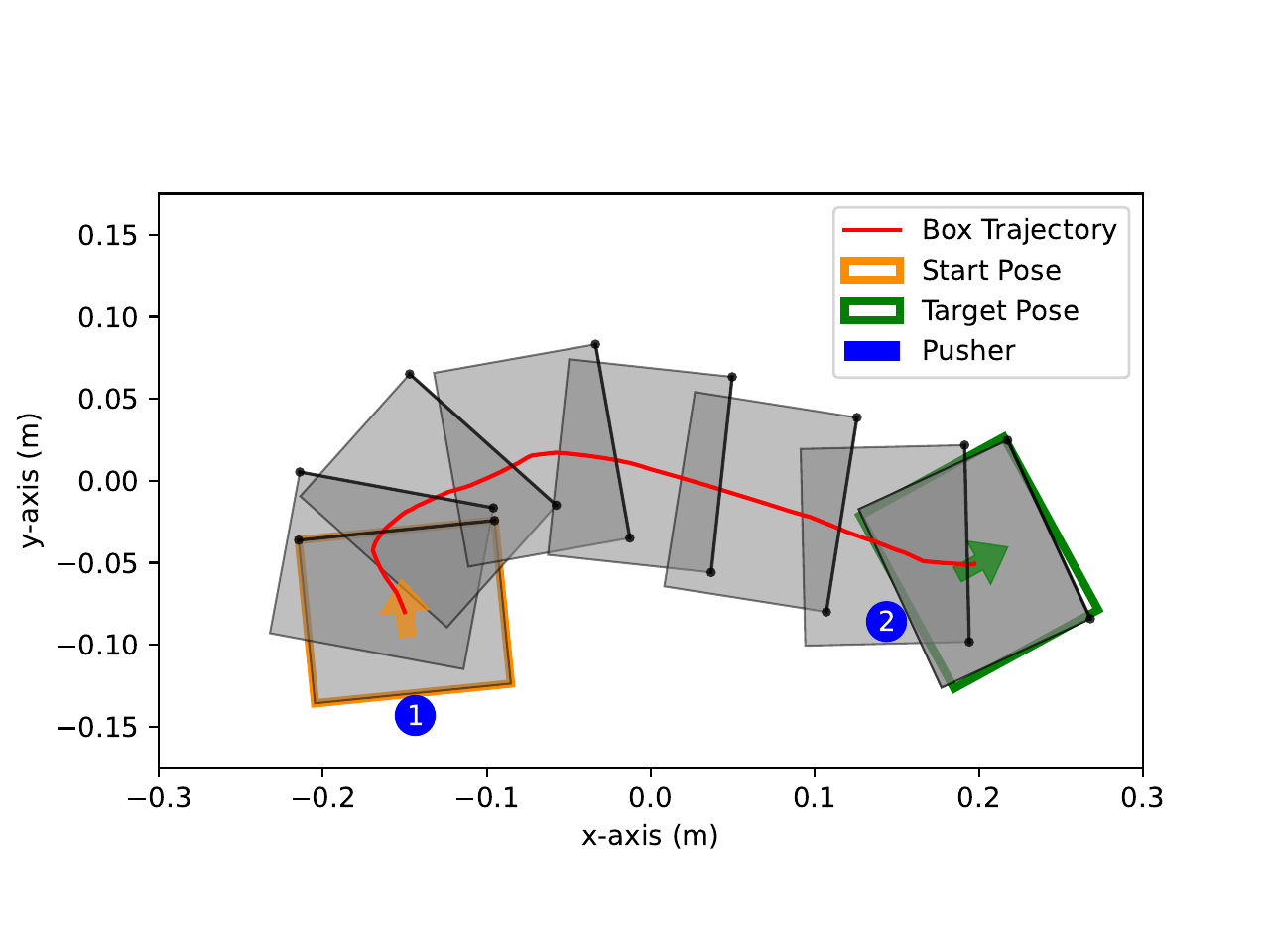}
        \caption{Simple pushing task}
        \label{fig:simple_trajectory}
    \end{subfigure}
    \begin{subfigure}{\linewidth}
        \centering
        \includegraphics[trim={0 1.2cm 0 2cm},clip, width=\linewidth]{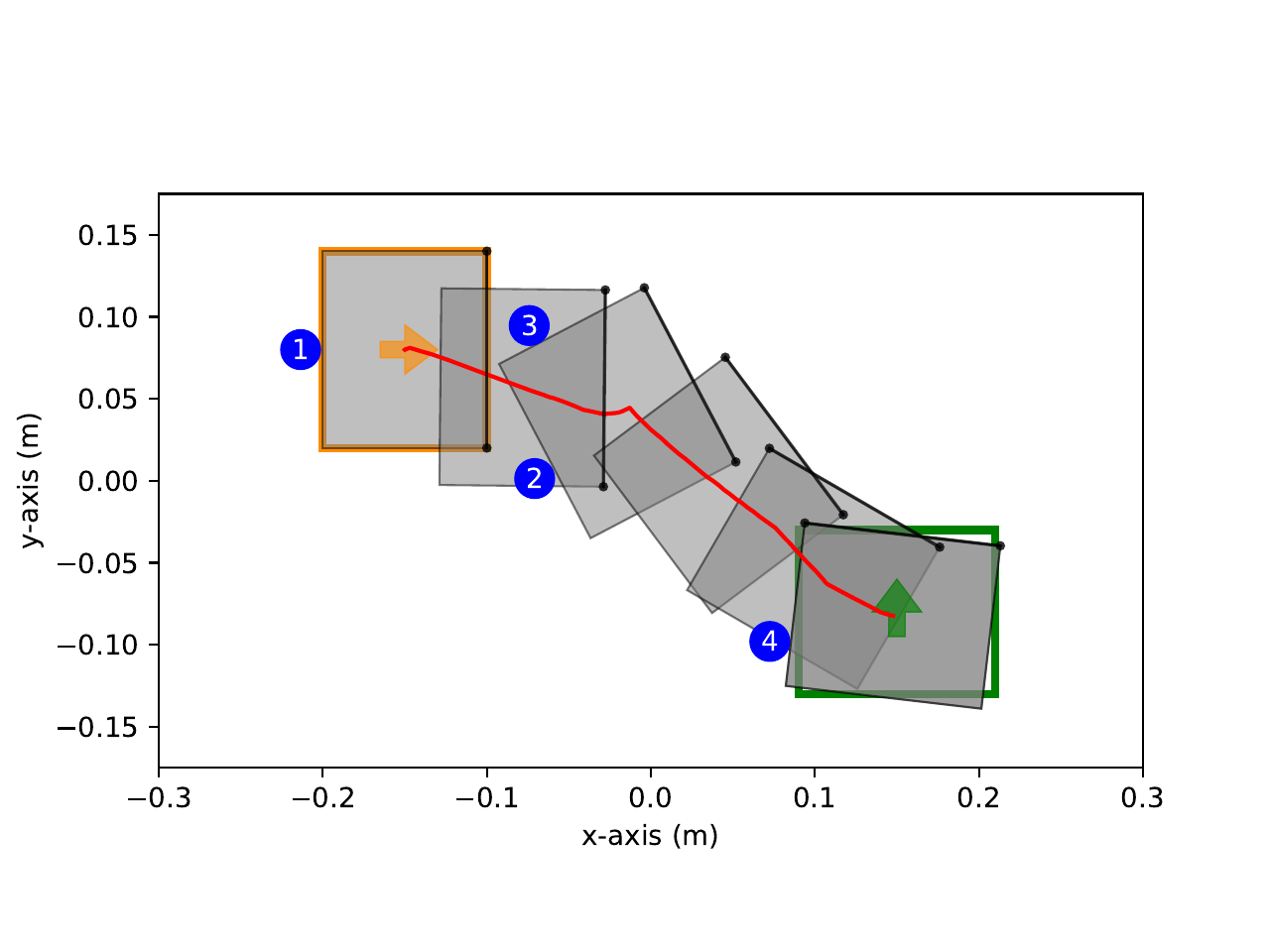}
        \caption{Pushing task with a face switch}
        \label{fig:trajectory_phase_switch}
    \end{subfigure}
    \caption{Trajectories generated by the PPO (LSTM + Categorical) policy in simulation. The pusher numbers indicate chronological order and the plot boundaries correspond to the workspace boundaries.}
\end{figure}

We also visualize the trajectories generated by the PPO (LSTM + Categorical) policy in simulation.
Fig.~\ref{fig:simple_trajectory} shows a standard pushing task and we see that the box follows a smooth and efficient trajectory, reaching the target pose with significant accuracy. 
The pusher remains in contact with the box throughout the entire trajectory.
Fig.~\ref{fig:trajectory_phase_switch} shows a more difficult pushing task where the policy performs a face switch.
The policy first re-orients the box, breaks contact at point 2, makes contact again at point 3, and finally pushes the box to the target.
Note that the boundary in the plot corresponds to the workspace boundary, so here a face switch seems necessary in order to solve the task.

\subsection{Robustness to Observation Noise}

In this experiment, we evaluate the robustness of the PPO (LSTM + Categorical) policy to different levels of correlated (per episode) and uncorrelated (per step) observation noise.
The problem with correlated noise is that it causes the observations of the environment to be consistently shifted, relative to the true state, throughout an episode.
On the other hand, moderate levels of uncorrelated noise pose fewer challenges since the noise update happens at every time step and it has an expected value of zero.
The results are recorded in Table~\ref{observation_noise}.
We report success rate and time to target, averaged over 1000 episodes with random starting configurations and target poses.
We increase the time limit from 10 to 30 seconds to allow the policy to make corrections if necessary.

We find that the policy is robust to both correlated and uncorrelated observation noise, achieving a good performance even in settings with large noise levels, which have standard deviations of comparable magnitudes to the success thresholds ($T_{x,y} = \SI{0.75}{\cm}$ and $T_\theta = \SI{0.17}{\radian} \approx 9.7$°).
As expected, correlated noise has a more significant effect on both the success rate and time to target.
Additionally, the policy seems to perform better with some uncorrelated noise rather than none.
This could be due to the uncorrelated noise causing the policy to receive more diverse observations of the environment, some of which could trigger certain policy behaviors that are beneficial for solving the task.

\begin{table}[t]
\caption{Success rate and time to target with correlated (per episode) and uncorrelated (per step) observation noise. Noise values correspond to the standard deviations of the sampling distributions. Results for the training set-up are underlined.}
\label{observation_noise}
\begin{center}
{\renewcommand{\arraystretch}{1.2}
\begin{tabular}{ll|llll}
\multicolumn{2}{l|}{\multirow{2}{*}{}}                                                                                                     & \multicolumn{4}{c}{\textbf{Uncorrelated Noise (SD)}}                                                                                                                                                                                                                          \\ \cline{3-6} 
\multicolumn{2}{l|}{}                                                                                                                      & \textit{\begin{tabular}[c]{@{}l@{}}0 cm\\ 0 rad\end{tabular}} & \textit{\begin{tabular}[c]{@{}l@{}}0.1 cm \\ 0.02 rad\end{tabular}} & \textit{\begin{tabular}[c]{@{}l@{}}0.3 cm\\ 0.06 rad\end{tabular}} & \textit{\begin{tabular}[c]{@{}l@{}}0.45 cm\\ 0.09 rad\end{tabular}} \\ \hline
\multicolumn{1}{l|}{\multirow{4}{*}{\rotatebox{90}{\textbf{Correlated Noise (SD)}}}} & \textit{\begin{tabular}[c]{@{}l@{}}0 cm\\ 0 rad\end{tabular}}       & \begin{tabular}[c]{@{}l@{}}99.1\%\\ 4.9 s\end{tabular}                 & \begin{tabular}[c]{@{}l@{}}99.7\%\\ 4.7 s\end{tabular}                       & \begin{tabular}[c]{@{}l@{}}99.4\%\\ 5.1 s\end{tabular}                      & \begin{tabular}[c]{@{}l@{}}99.0\%\\ 5.4 s\end{tabular}                      \\ \cline{2-6} 
\multicolumn{1}{l|}{}                                                & \textit{\begin{tabular}[c]{@{}l@{}}0.1 cm \\ 0.02 rad\end{tabular}} & \begin{tabular}[c]{@{}l@{}}99.0\%\\ 4.8 s\end{tabular}                 & \begin{tabular}[c]{@{}l@{}} \underline{99.2}\%\\ \underline{5.0 s}\end{tabular}                       & \begin{tabular}[c]{@{}l@{}}99.0\%\\5.1 s\end{tabular}                      & \begin{tabular}[c]{@{}l@{}}99.2\%\\ 5.4 s\end{tabular}                      \\ \cline{2-6} 
\multicolumn{1}{l|}{}                                                & \textit{\begin{tabular}[c]{@{}l@{}}0.3 cm\\ 0.06 rad\end{tabular}}  & \begin{tabular}[c]{@{}l@{}}95.7\%\\ 6.2 s\end{tabular}                 & \begin{tabular}[c]{@{}l@{}}97.0\%\\ 6.0 s\end{tabular}                       & \begin{tabular}[c]{@{}l@{}}97.2\%\\ 6.4 s\end{tabular}                      & \begin{tabular}[c]{@{}l@{}}96.4\%\\ 6.8 s\end{tabular}                      \\ \cline{2-6} 
\multicolumn{1}{l|}{}                                                & \textit{\begin{tabular}[c]{@{}l@{}}0.45 cm\\ 0.09 rad\end{tabular}}  & \begin{tabular}[c]{@{}l@{}}84.5\%\\ 9.1 s\end{tabular}                 & \begin{tabular}[c]{@{}l@{}}87.5\%\\ 8.8 s\end{tabular}                       & \begin{tabular}[c]{@{}l@{}}91.3\%\\ 8.3 s\end{tabular}                      & \begin{tabular}[c]{@{}l@{}}90.6\%\\ 9.0 s\end{tabular}                     
\end{tabular}
}
\end{center}
\end{table}

\subsection{Limits of Gaussian Exploration}

\begin{figure}[t]
    \centering
    \includegraphics[trim={0 0.2cm 0 1.15cm},clip,width=\linewidth]{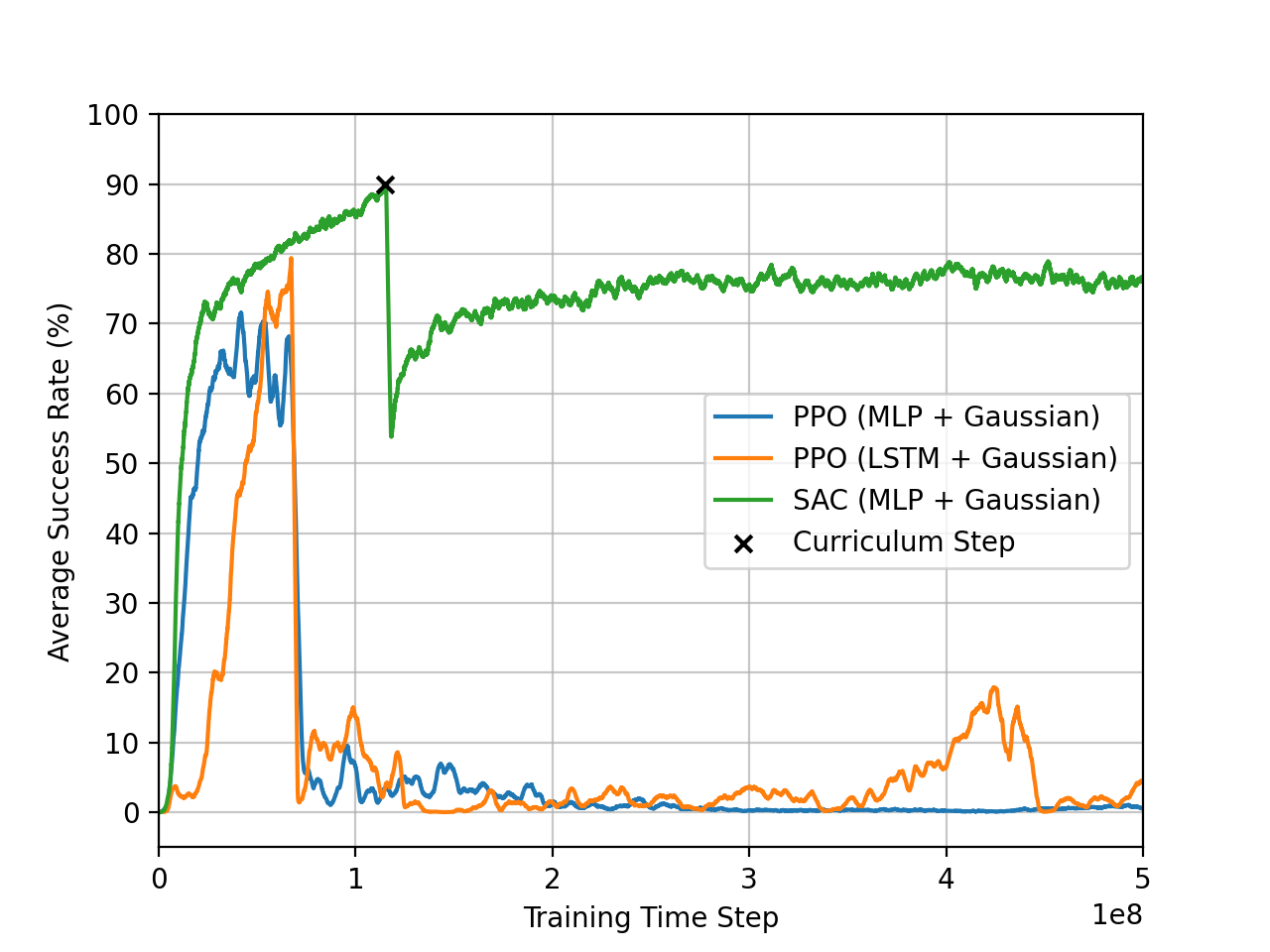}
    \caption{Training performance of the policies with Gaussian exploration on a simplified planar pushing task. A curriculum progressively increases the range of starting and target orientations at 90\% average success rate.}
    \label{fig:gaussian_simple_task}
\end{figure}

In order to better understand why the policies with Gaussian exploration failed to learn the planar pushing task, we examine their performance in a simplified setting.
We limit the starting position of the box to the left side of the workspace, and the target position to the right side of the workspace. 
Additionally, we restrict the sampling distribution for the starting and target orientations of the box to $\mathcal{U}([-\pi/4, \pi/4])$ \SI{}{\radian}, and remove the synthetic external disturbances.
The starting position of the pusher is always in the backside of the box, from where it can naturally push it towards the target in the right side of the workspace. 
For this experiment, we use success thresholds $T_{x,y} = \SI{1.5}{\cm}$ and $T_\theta = \SI{0.34}{\radian} \approx 19.5$° throughout.
Furthermore, we define a curriculum for training such that, if the policy reaches a 90\% average success rate, the width of the uniform sampling distribution for the starting and target orientations increases by $\pi/2$, still remaining centered around 0.
The resulting learning curves are shown in Fig.~\ref{fig:gaussian_simple_task}.

We can see that the training performance of the PPO policies, besides remaining below 80\% success rate, collapses early on.
The SAC policy reaches 90\% success rate once, which triggers the first step in the curriculum, corresponding to a range of starting and target orientations of $[-\pi/2, \pi/2]$ \SI{}{\radian}; however, it fails to reach the next curriculum step.
In simplified settings, with small starting and target orientations, the pusher might be able to achieve the task without breaking contact, unlike for larger orientations which will require sharper turns and hence face switching.
Therefore, the reason why these policies fail to progress with the curriculum could be that they struggle with the tasks that require face switching since their exploration strategy is unable to effectively capture this additional modality.

\begin{figure*}[t]
    \begin{subfigure}{0.16\linewidth}
        \centering
        \includegraphics[width=\linewidth]{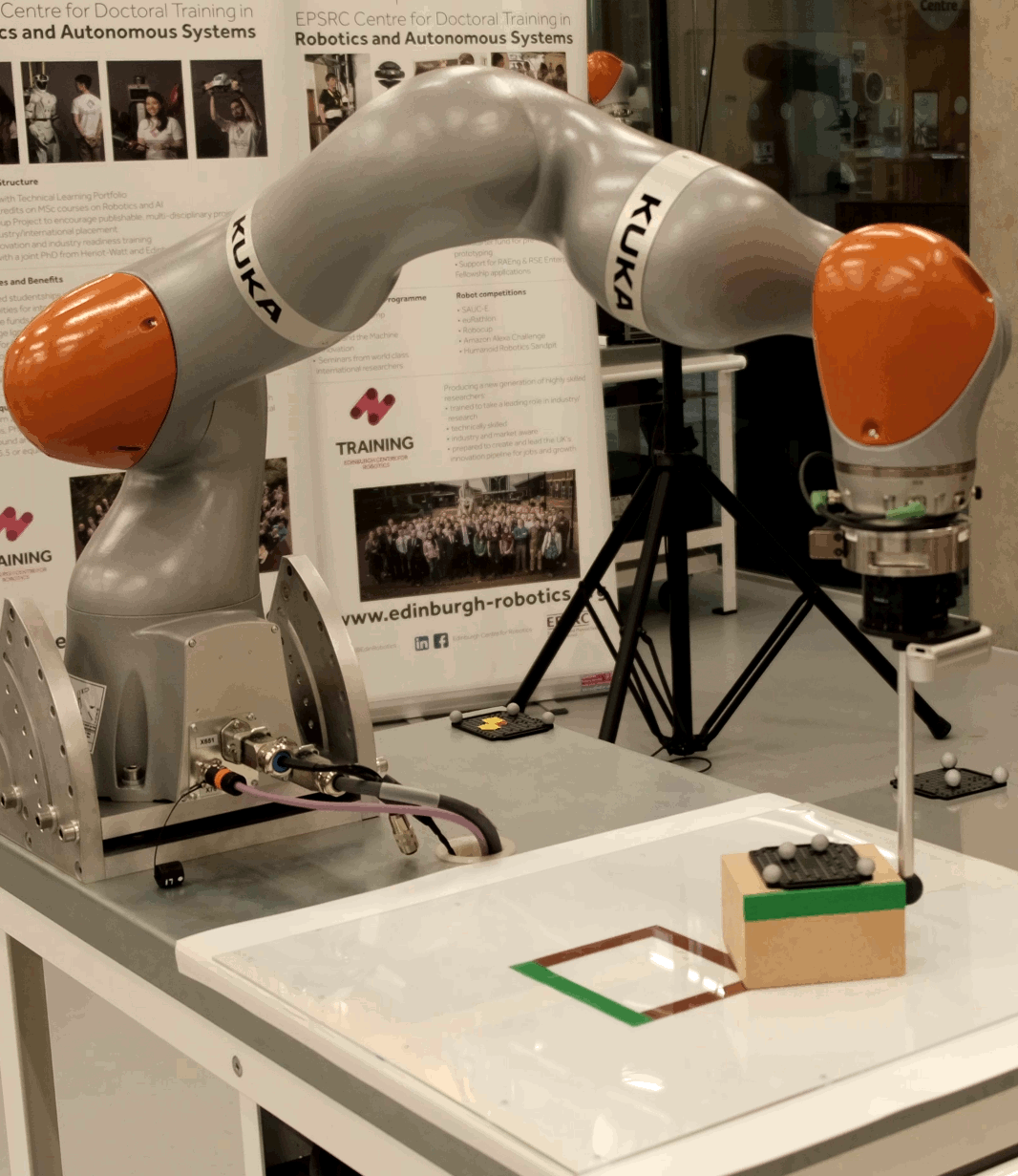}
        \caption{}
        \label{fig:key_frame_1}
    \end{subfigure}
    \begin{subfigure}{0.16\linewidth}
        \centering
        \includegraphics[width=\linewidth]{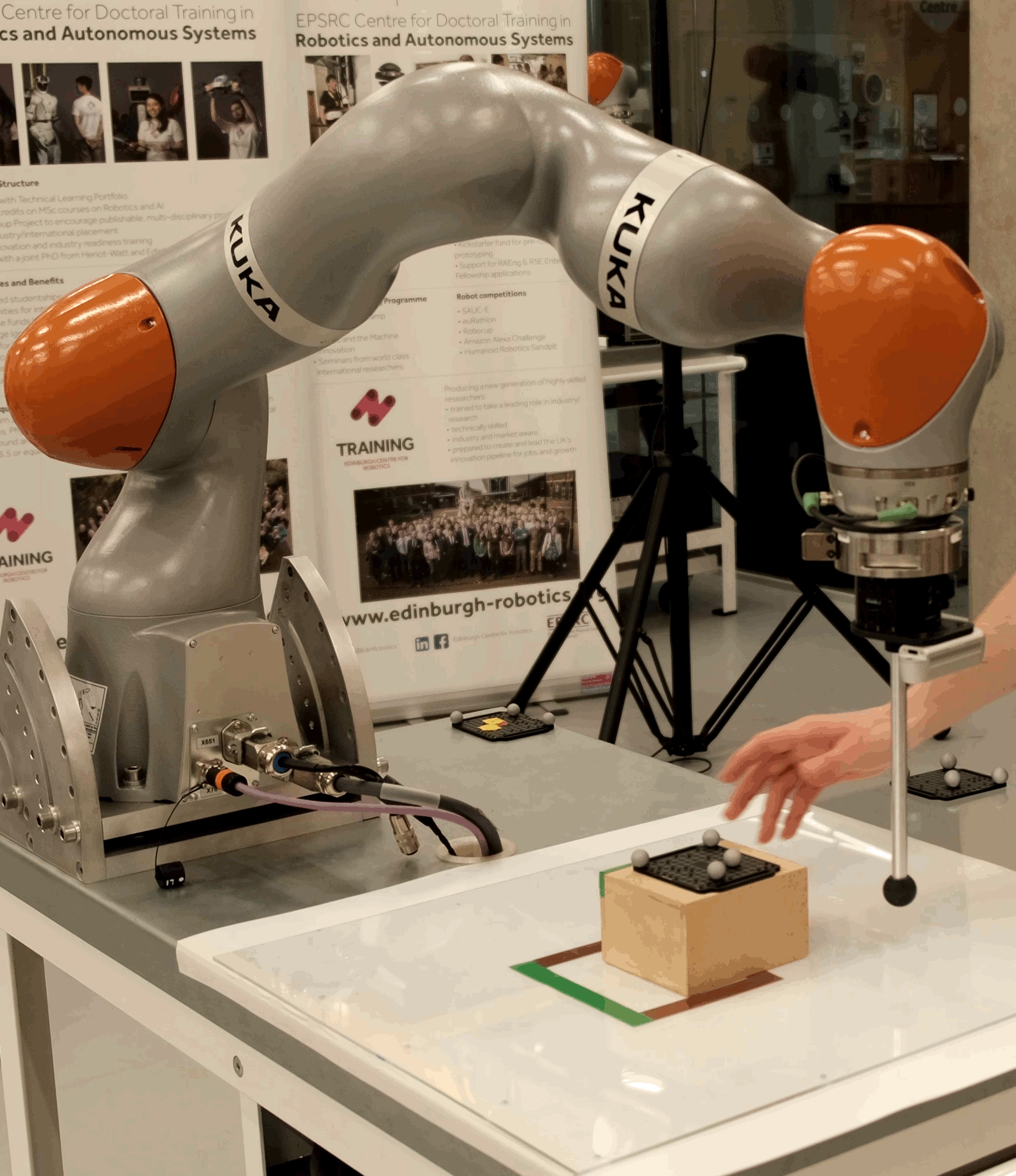}
        \caption{}
        \label{fig:key_frame_2}
    \end{subfigure}
    \begin{subfigure}{0.16\linewidth}
        \centering
        \includegraphics[width=\linewidth]{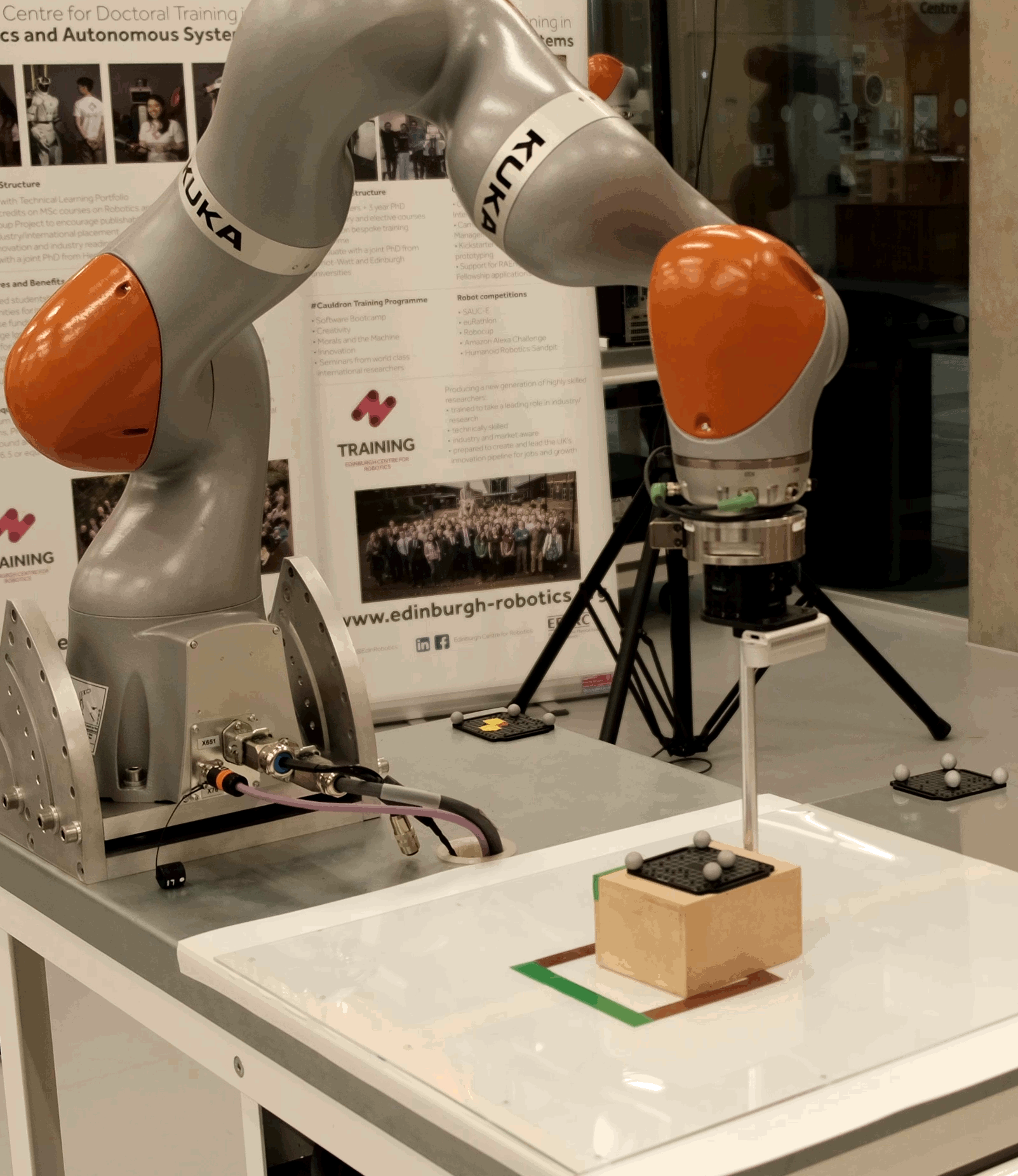}
        \caption{}
        \label{fig:key_frame_3}
    \end{subfigure}
    \begin{subfigure}{0.16\linewidth}
        \centering
        \includegraphics[width=\linewidth]{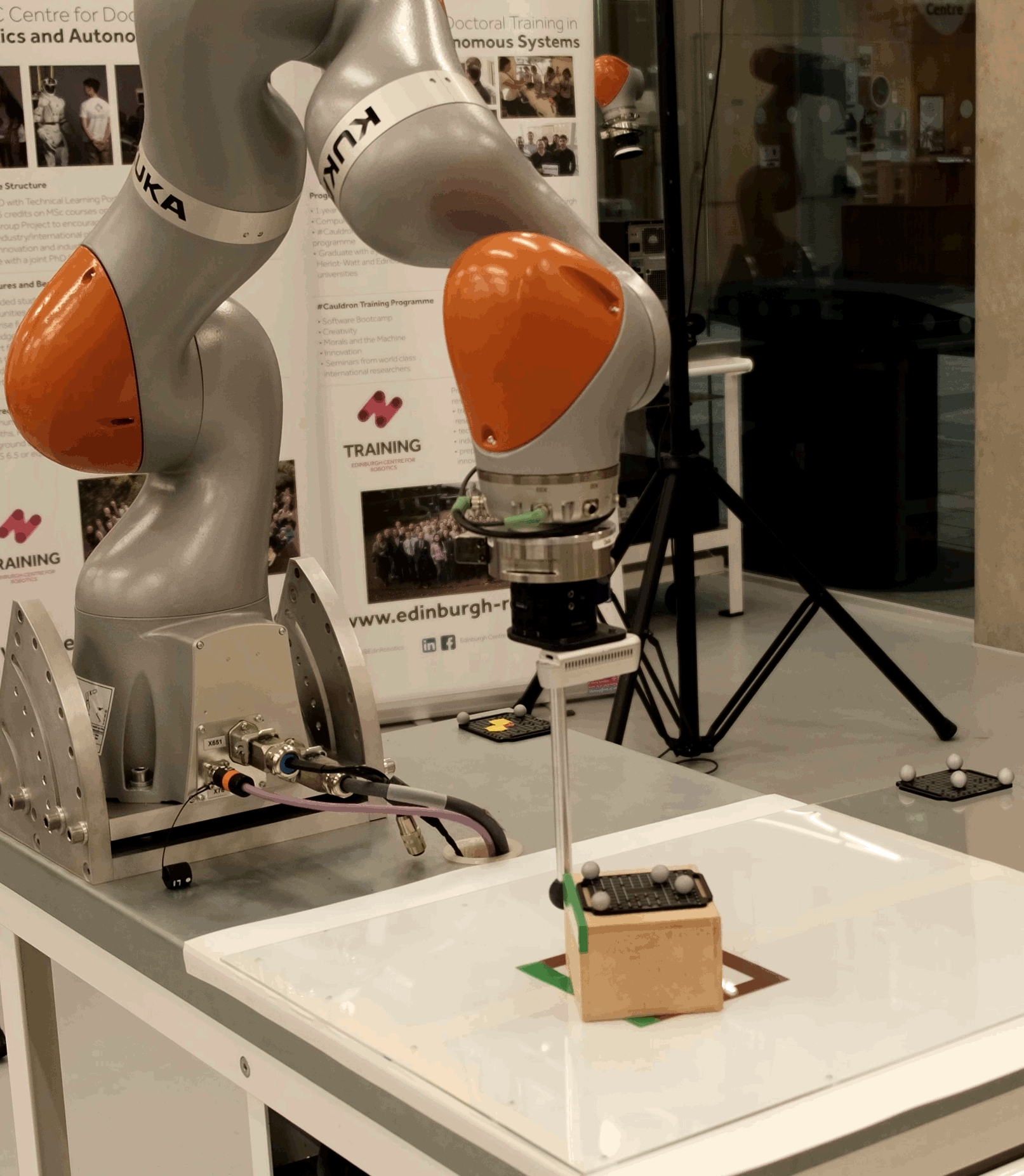}
        \caption{}
        \label{fig:key_frame_4}
    \end{subfigure}
    \begin{subfigure}{0.16\linewidth}
        \centering
        \includegraphics[width=\linewidth]{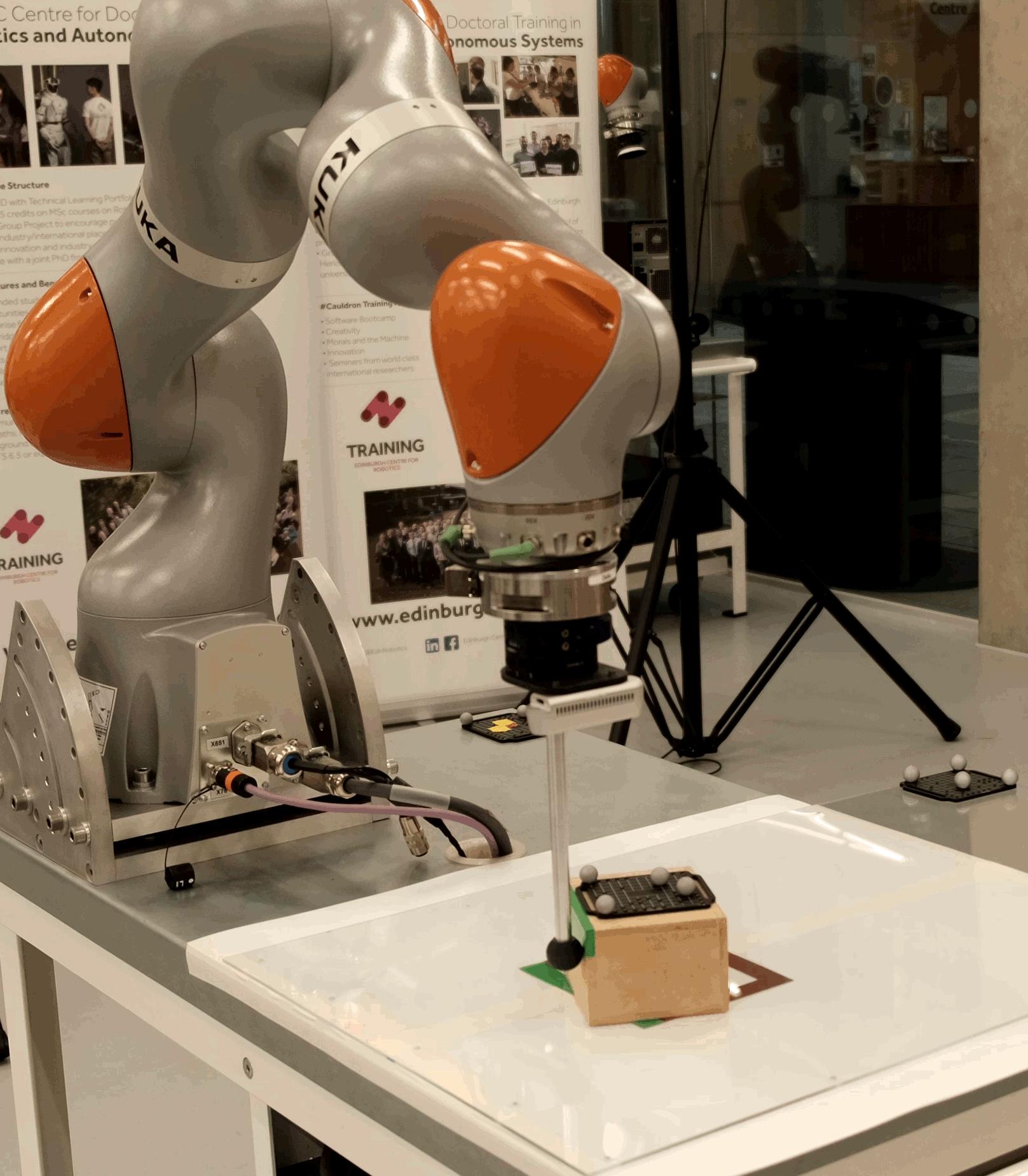}
        \caption{}
        \label{fig:key_frame_5}
    \end{subfigure}
    \begin{subfigure}{0.16\linewidth}
        \centering
        \includegraphics[width=\linewidth]{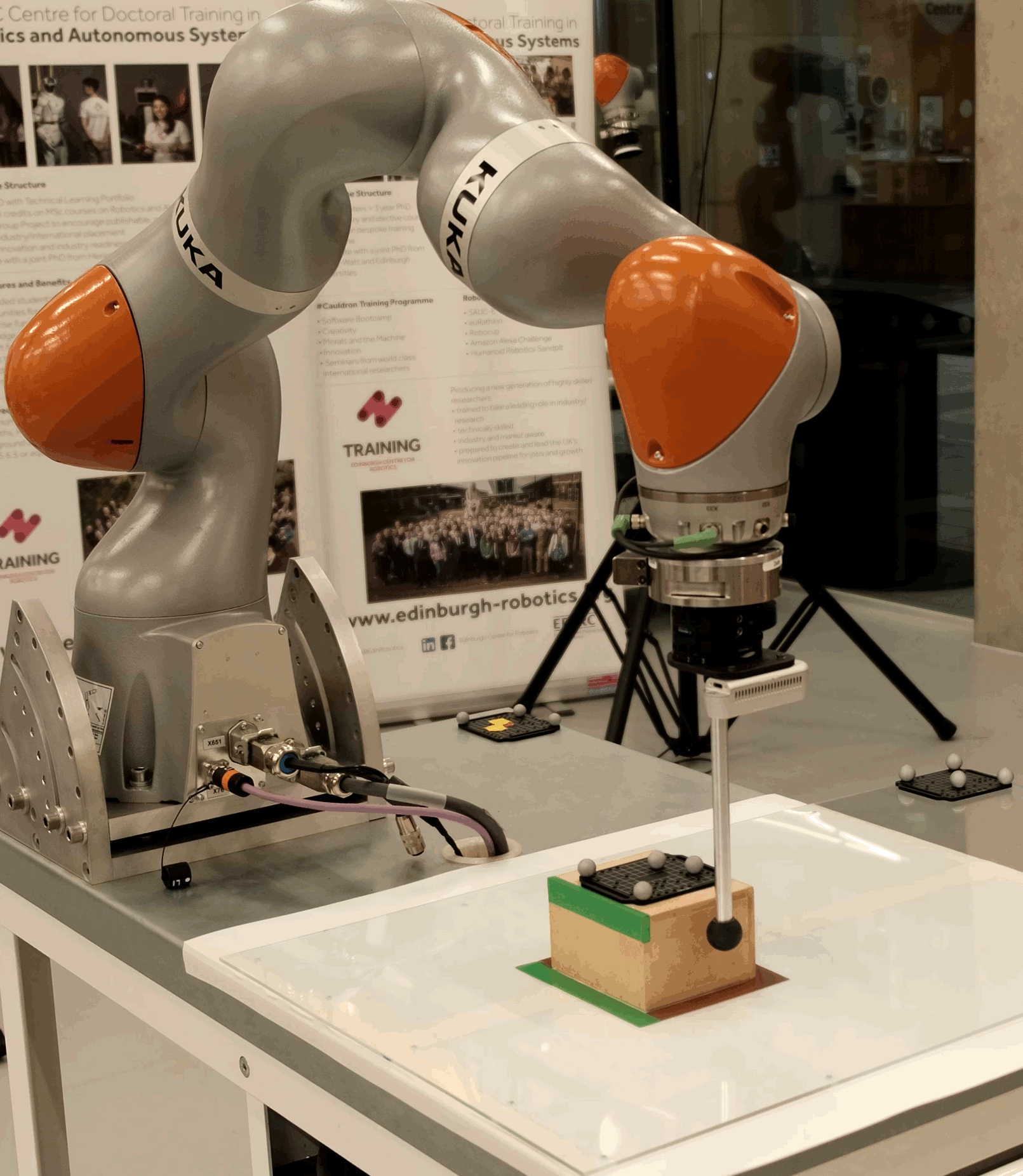}
        \caption{}
        \label{fig:key_frame_6}
    \end{subfigure}
  \caption{%
  Key frames of the KUKA iiwa robot pushing the box to a target pose.
  (a) Shows the starting configuration, a large disturbance is applied in (b), and (c)-(f) exhibit the RL policy recovering from the disturbance and reaching the goal.
  }
  \label{fig:kuka_key_frames}
\end{figure*}

\subsection{Scalability with Two Pushers}

\begin{figure}[b]
    \centering
    \includegraphics[trim={0 0.2cm 0 1.15cm},clip,width=\linewidth]{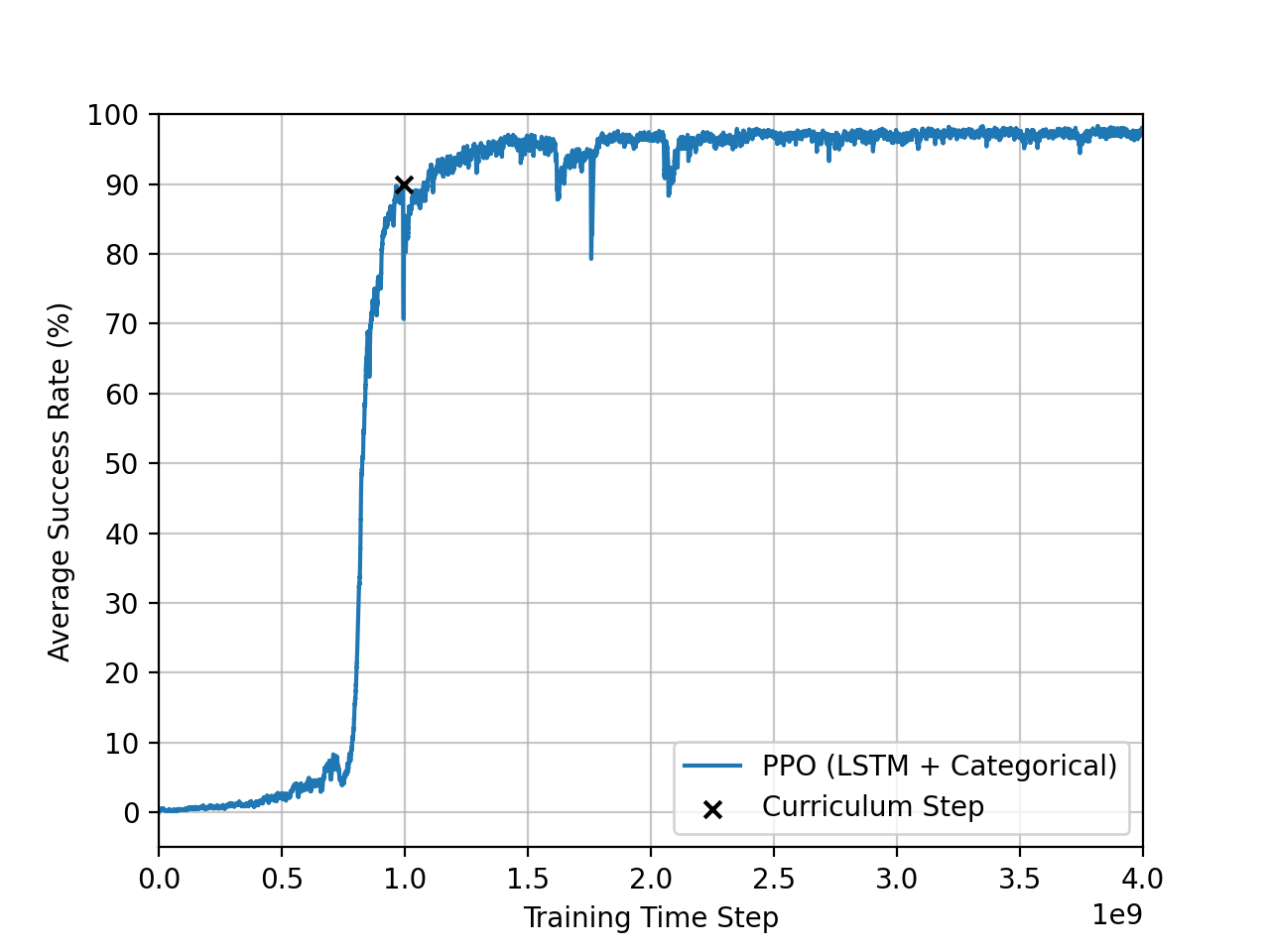}
    \caption{Training performance of the PPO (LSTM + Categorical) policy on a planar pushing environment with two pushers.}
    \label{fig:training_curves_two_pushers}
\end{figure}

To evaluate the scalability of our framework, we train the PPO (LSTM + Categorical) policy on a planar pushing task with two pushers. 
For this experiment, we use our original curriculum, as described in \cref{sec:curriculum}, with the same success thresholds as in the one-pusher set-up.
The results are shown in Fig. \ref{fig:training_curves_two_pushers}.
The policy scales well to this more complex task, and manages to achieve a success rate greater than $97\%$ with the reduced success thresholds. 
Nevertheless, convergence is slower, which is expected due to the increased dimensionality of the problem. 

\section{Experiments and Results -- Hardware}

\begin{figure}[b]
    \centering
    \includegraphics[trim={0 1.2cm 0 2cm},clip,width=\linewidth]{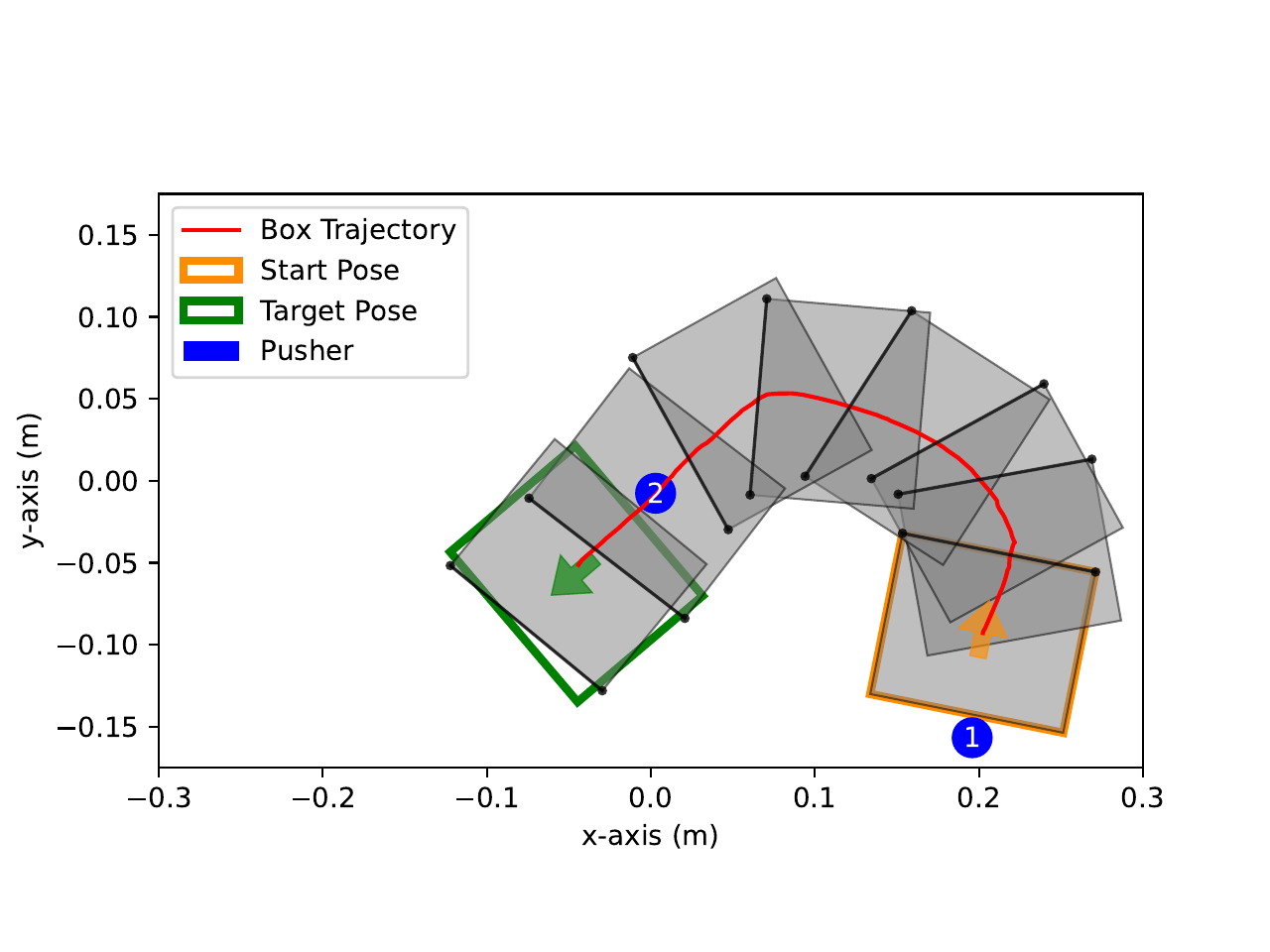}
    \caption{Trajectory generated by the PPO (LSTM + Categorical) policy in the physical robot.}
    \label{fig:trajectory_real_robot}
\end{figure}

We investigate the performance of the PPO (LSTM + Categorical) policy on a physical planar pushing hardware set-up with the KUKA iiwa robot.
We use the Vicon motion capture system to track the current and target box pose, and use OpTaS~\cite{Mower2023} to map the policy actions in the end-effector task-space to the robot joint configurations.
We use the ROS-Pybullet interface~\cite{Mower2022} to develop and test the robot software implementation.

\cref{fig:kuka_key_frames} shows a sequence of key frames of the robot pushing the box to a target pose and recovering from an external disturbance.
The supplemental video (\href{https://youtu.be/vTdva1mgrk4}{\textcolor{urlcolor}{https://youtu.be/vTdva1mgrk4}}) clearly demonstrates the behavior of the trained policy.
We find that the policy translates well to the real world and is able to effectively cope with the dynamics of the new environment. 
Additionally, the policy is robust to large external disturbances and changes in the target pose. 
\cref{fig:trajectory_real_robot} shows a sample trajectory generated by the robot.
The policy manages to remain within the workspace boundaries and generates a smooth trajectory towards the target box pose. 

We also obtain statistics for the average success rate and time to target of the policy on the robot.
We use success thresholds $T_{x,y} = \SI{0.75}{\cm}$ and $T_\theta = \SI{0.17}{\radian} \approx 9.7$°, and enforce a time limit of 30 seconds. 
To collect the data, we generate 5 random target box poses and, for each target pose, we run the policy from 15 random starting configurations. 
The resulting average success rate is 97.3\% and the average time to target is 6.5 seconds.
The policy exhibits similar performance to simulation, which indicates good transferability to the physical hardware.

\section{Summary and Discussion}

In this paper, we have proposed a multimodal exploration approach, through categorical distributions on a discrete action space, to enable the learning of planar pushing RL policies for arbitrary initial and target object poses, i.e.~different positions and orientations, with improved accuracy.
Our experiments demonstrate that the learned policies are robust to observation noise and external disturbances, produce smooth trajectories, and scale well to two pushers.
Furthermore, we have validated that the policy, trained only in simulation, achieves both smooth motions and small target error when executed on the physical robotic hardware.

One of the key realizations in this work was that, when attempting to learn planar pushing RL policies for the case of arbitrary object orientations, the use of a multivariate Gaussian with diagonal covariance for exploration as per previous literature~\cite{peng_2018, lowrey_2018, cong_2022, jeong_2019}, would lead to the RL failing to converge.
Borrowing the insight from the model-based literature~\cite{hogan_2020,moura_2022}, that planar pushing has hybrid-dynamics reflected in a set of different contact modes that constraint the control actions, we hypothesized that we can reason about planar pushing as a multimodal control problem.
Therefore, we proposed describing the action space through categorical distributions to capture the multimodal nature of the problem, potentially leading to more effective exploration of different contact modes during training.
We have also shown that indeed, during training, the categorical action distributions exhibit multimodal exploration strategies. 

For future work, we will continue studying multimodal exploration strategies, aiming to retain the continuity of the action space.
Furthermore, we will explore implicit representations of the learned policy, such as in \cite{florence_2022}, as another way to capture the multimodal and discontinuous nature of the pushing task.
We also intend to extend this framework to more complex manipulation tasks with previously unseen object geometry and incorporating onboard perception.

\printbibliography

\end{document}